\documentclass[10pt,twocolumn,letterpaper]{article}
\usepackage[pagenumbers]{cvpr}
\usepackage{amsmath,amssymb}
\usepackage{booktabs}
\usepackage{tabularx}
\usepackage{array}
\usepackage{colortbl}
\usepackage{microtype}
\usepackage{algorithm}
\usepackage{algpseudocode}
\definecolor{cvprblue}{rgb}{0.21,0.49,0.74}
\definecolor{oursbg}{HTML}{EAF1FA}
\definecolor{deltagray}{HTML}{6B6B66}
\usepackage[pagebackref,breaklinks,colorlinks,allcolors=cvprblue]{hyperref}
\hypersetup{
    pdftitle={LRQ-Cache: Compressing Feature Caches in Video Diffusion via Low-Rank Approximation and Quantization},
    pdfauthor={Ruizhe Yao, Zhen Huang, Songpan Gao, Danyi Liu, Zijian Cao, Weijie Zheng, Haohuan Fu},
    pdfsubject={Feature-cache compression via low-rank approximation and factor quantization},
    pdfkeywords={Video Diffusion Models; Feature Cache Compression; Low-Rank Approximation; Factor Quantization; Temporal Reuse}
}
\def\confName{CVPR}
\def\confYear{2027}
\title{LRQ-Cache: Compressing Feature Caches in Video Diffusion\\
via Low-Rank Approximation and Quantization}
\author{Ruizhe Yao$^{1,*}$ \quad Zhen Huang$^{2,*}$ \quad Songpan Gao$^{1,3}$ \quad Danyi Liu$^{2}$\\
Zijian Cao$^{4}$ \quad Weijie Zheng$^{1,\dagger}$ \quad Haohuan Fu$^{2,\dagger}$\\[0.6em]
$^{1}$Harbin Institute of Technology, Shenzhen \quad $^{2}$Tsinghua University\\
$^{3}$Peng Cheng Laboratory \quad $^{4}$Northeastern University\\[0.4em]
{\small $^{*}$Equal contribution. \quad $^{\dagger}$Corresponding authors.}}
\begin{document}
\maketitle

\begin{abstract}
Video diffusion transformers (DiTs) achieve high-fidelity generation, but inference is expensive because every denoising step evaluates a large transformer over all spatiotemporal tokens of the video. Temporal caching accelerates them by reusing intermediate features across denoising steps.
However, retaining dense feature tensors introduces a persistent storage cost that grows with video length, resolution, and cache coverage, reaching 109--346\,GiB per video for the predictive caches we evaluate.
Such states exceed the memory of a single GPU, so native caches must either be offloaded to the CPU, which slows generation, or be sharded across several GPUs, e.g., with sequence parallelism, so that the cache rather than the compute sets the number of GPUs.
We instead shrink the cached state itself, exploiting its redundancy: a few channel patterns shared across tokens capture much of the energy of each cached feature.
Based on this observation, we propose \textbf{LRQ-Cache}, a training-free method that combines low-rank approximation and quantization for feature-cache compression.
LRQ-Cache represents each cached feature with a small set of factors, quantizes them before storage, and reconstructs the feature at reuse time, leaving cache mechanisms unchanged.
Across three caching mechanisms on HunyuanVideo and Wan2.1, LRQ-Cache reduces the persistent
state of predictive caches by 42--71$\times$, to 2.3--4.9\,GiB per video.
VBench stays within 0.25\,pp of the native cache for TeaCache, and the larger drops of the Wan2.1 predictive caches, up to 3.5\,pp, shrink at higher ranks.
As a result, predictive caches that exceed the memory of a single 80\,GB H100 GPU fit on it with LRQ-Cache, making generation 2.7--5.9$\times$ faster than with CPU-offloaded native caches.
\end{abstract}

\section{Introduction}
\label{sec:intro}
Diffusion transformers (DiTs)~\cite{peebles2023scalable} have become the dominant architecture for high-fidelity video generation~\cite{yang2025cogvideox,kong2024hunyuanvideo,wan2025wan}.
However, their inference remains expensive: a video is generated through tens of denoising steps, and each step evaluates a large transformer over all spatiotemporal tokens of a latent video, for example 118,800 tokens for a 129-frame 720p video in HunyuanVideo.
This cost is largely repetitive: prior work has observed that features change only slightly between adjacent denoising steps and evolve slowly and continuously over the trajectory~\cite{ma2024deepcache,selvaraju2024fora,liu2025taylorseer}, which makes temporal redundancy a natural source of inference savings.

Feature caching turns this redundancy into speed: it computes the network in full only at refresh steps and reuses the stored results at the steps in between.
Residual caches such as TeaCache add a stored whole-stack residual to the current input~\cite{liu2025teacache}.
Because features drift as reuse extends farther from a refresh, predictive caches go one step further and forecast later features; the TaylorSeer and ScalingCache configurations we evaluate retain outputs and temporal differences for each cached attention or MLP module to forecast its later features~\cite{liu2025taylorseer,gu2026scalingcache}.

Reuse, however, requires keeping these features between refreshes, and because they span all video tokens, their storage grows with video length, resolution, and the number of cached modules.
Predictive caches such as TaylorSeer and ScalingCache are affected most, since they keep an output and a finite difference for every attention and MLP module of every layer: their state reaches 109\,GiB per video on HunyuanVideo (129 frames at 720p), and 150 and 346\,GiB on Wan2.1 T2V and I2V (81 frames at 480p and 720p).
These states exceed the memory of a single 80\,GB GPU, so native predictive caches must either be offloaded to the CPU, where transfers take 70--89\% of the generation time in our measurements, or be sharded across several GPUs, e.g., with sequence parallelism, so that the cache rather than the compute sets the number of GPUs.

\begin{figure*}[t]
    \centering
    \includegraphics[width=\textwidth]{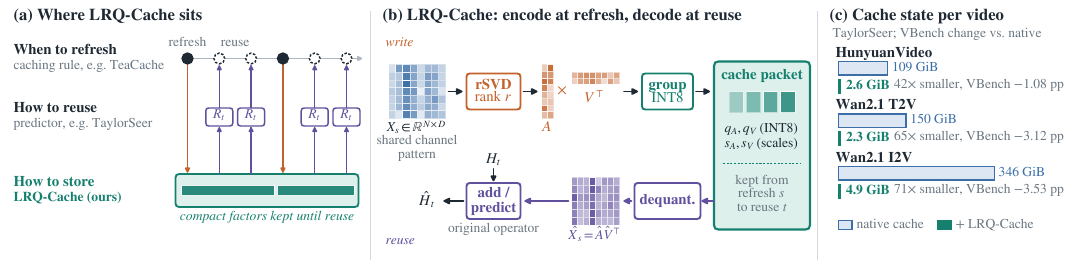}
    \caption{\textbf{Overview of LRQ-Cache.}
    (a) Feature caching computes the network at refresh steps and reuses the stored state at the steps in between: the caching rule decides when to refresh, and the predictor $R_t$ decides how to reuse.
    LRQ-Cache leaves both unchanged and determines how the retained state is stored.
    (b) At a refresh step, LRQ-Cache factorizes the cached tensor $X_s$ with randomized SVD and stores the factors as group-wise INT8 codes; at a reuse step, it dequantizes the factors and reconstructs $\hat X_s=\hat A\hat V^{\top}$ for the original add or predict operator.
    (c) Persistent TaylorSeer state per video with native caching and with LRQ-Cache, and the VBench change of LRQ-Cache relative to the native cache.}
    \label{fig:overview}
\end{figure*}

Existing work has largely left this storage untouched and instead improves when and how cached features are reused (Fig.~\ref{fig:overview}(a)).
Caching rules select refresh steps~\cite{liu2025teacache,gu2026scalingcache}, while reuse operators extrapolate or revise cached estimates~\cite{liu2025taylorseer,gu2026scalingcache,lan2026postcache}.
Other work changes what is cached~\cite{liu2025freqca,cui2026predict,lu2026barycache} or where cache operators run~\cite{zhang2026coda}.
This leaves open whether the cached state itself can be compressed while retaining the cache sites and the definitions of the underlying caching mechanism's reuse operators.

To answer this question, we examine what a cached feature contains.
Much of it is shared across tokens: for the HunyuanVideo TeaCache residual, a single channel pattern accounts for a median 91.4\% of the energy over two evaluation inputs and three refreshes per input.
Such redundancy across tokens suggests that the feature need not be stored densely.
Concretely, repeating one pattern at every token forms a rank-one matrix, and additional components can describe how individual tokens deviate from it, so the feature can be stored approximately as low-rank factors instead of a dense tensor.
The structure is not equally strong everywhere, however: in predictive caches, it holds for the outer layers but weakens in the middle layers and in finite differences.
Building on this redundancy, we propose \textbf{LRQ-Cache}, a training-free storage interface for feature caching.
At refreshes, it factorizes each cached video feature online and stores the factors as INT8 codes with scales and offsets; at reuse, it reconstructs the feature for the reuse operator.
Because it changes only how cached tensors are stored, the model weights, cache sites, refresh rules, and reuse operators of the underlying caching mechanism stay unchanged, and one codec serves both residual and predictive caches.

To test whether this compact form preserves quality and removes the memory bottleneck, we apply LRQ-Cache with a fixed rank of 128 to TeaCache, TaylorSeer, and ScalingCache on HunyuanVideo and Wan2.1 T2V/I2V, over 50 inputs per model and task.
It shrinks the state of predictive caches by 42--71$\times$, from 109--346\,GiB to 2.3--4.9\,GiB per video (Table~\ref{tab:main}).
VBench stays within 0.25\,pp of the native cache for TeaCache and drops by at most 3.5\,pp for predictive caches, a loss that shrinks at higher ranks.
As a result, the compressed predictive caches of all three models fit on a single 80\,GB GPU, making generation 2.7--5.9$\times$ faster than with CPU-offloaded native caches.

In summary, our contributions are as follows:
\begin{itemize}
    \item \textbf{The cached state as a bottleneck.} We identify the stored state of feature caches, rather than their computation, as what keeps predictive caching of long, high-resolution videos from running on a single GPU, and show that cached features are redundant across tokens: a single shared channel pattern carries 91.4\% of the energy of the TeaCache residual.
    \item \textbf{LRQ-Cache.} We propose a training-free storage interface that keeps cached features as INT8-quantized low-rank factors fitted online and reconstructs them at reuse, with a single rank setting its memory--fidelity trade-off. To our knowledge, it is the first method that compresses the features stored by existing step-wise feature caches of video DiTs while leaving the model, its computation, and the caching mechanism unchanged; prior work lowers cache memory by caching fewer features~\cite{liu2025freqca,lu2026barycache} or by quantizing the model~\cite{liu2025cachequant}.
    \item \textbf{Evaluation across models and mechanisms.} On three video DiTs and three caching mechanisms, LRQ-Cache shrinks the state of predictive caches by 42--71$\times$, to 2.3--4.9\,GiB per video, so that they run on a single GPU 2.7--5.9$\times$ faster than with CPU-offloaded native caches. We further show that the reuse operator of predictive caches amplifies the remaining reconstruction error, which explains where quality is lost.
\end{itemize}

\section{Related Work}
\label{sec:related}
LRQ-Cache draws on three lines of work.
Acceleration methods reduce the computation of video DiTs, feature caching reuses this computation across denoising steps at the cost of stored features, and compressed representations of cached activations reduce such storage.
We review them in this order and relate each to LRQ-Cache.

\paragraph{Efficient inference of video DiTs.}
DiTs operate on latent patches~\cite{peebles2023scalable}, and video generators extend them to spatiotemporal tokens~\cite{yang2025cogvideox,kong2024hunyuanvideo,wan2025wan}, so the cost of each denoising step grows with video length and resolution.
Step distillation and consistency models train the model to generate in fewer steps~\cite{salimans2022progressive,song2023consistency}, whereas sparse attention restricts computation to a subset of token pairs~\cite{xi2025sparsevideogen,yang2025sparsevideogen2} and quantization lowers the precision of weights and activations~\cite{li2025svdquant,su2026sixbit}.
These methods change how each step is computed.
Feature caching instead skips part of the computation at some steps by reusing features from earlier ones, which requires storing those features; LRQ-Cache targets this storage and leaves the computation of full steps unchanged.

\paragraph{Feature caching and its memory cost.}
Early caches reuse the outputs of whole network blocks across adjacent steps~\cite{ma2024deepcache,selvaraju2024fora,ma2024l2c}.
Later work refines when and where to reuse by selecting tokens~\cite{zou2025toca}, distinguishing attention types~\cite{zhao2025pab}, adapting refresh decisions to timesteps and video content~\cite{liu2025teacache,kahatapitiya2025adacache}, or extending reuse to classifier-free guidance~\cite{lyu2025fastercache}.
Others improve how cached features are reused: TaylorSeer~\cite{liu2025taylorseer}, ScalingCache~\cite{gu2026scalingcache}, and BaryCache~\cite{lu2026barycache} extrapolate them by Taylor expansion, scaled differences, and barycentric weights, respectively; PrediT forecasts outputs from their history~\cite{cui2026predict}, PoSTCache revises earlier estimates by postdictive smoothing~\cite{lan2026postcache}, and DisCa trains a lightweight predictor~\cite{zou2026disca}.
Further methods change what is cached, such as a whole-stack residual~\cite{liu2025teacache} or a cumulative residual feature~\cite{liu2025freqca}.
Low-rank structure has so far served estimation: SVD-Cache forecasts the principal component of a feature~\cite{chen2026svdcache}, and increment-calibrated caching corrects reused outputs with low-rank approximations of pretrained weights~\cite{chen2025icc}.
Few methods address the memory of the cache itself: CacheQuant quantizes weights and activations jointly with caching but keeps the token--channel layout~\cite{liu2025cachequant}, and CODA moves memory-bound cache operators to a near-memory engine~\cite{zhang2026coda}.
LRQ-Cache instead uses low rank for storage: it changes only how cached tensors are represented, keeping the cache sites and reuse operators of the underlying mechanism, without training.

\paragraph{Compressing cached activations.}
Storing a cache in compact form and reconstructing it when read has been studied mainly for the key--value cache of language models, through low-bit quantization~\cite{liu2024kivi}, low-rank projections obtained offline from the projection weights~\cite{chang2025palu}, or quantization with low-rank and sparse error correction~\cite{kang2024gear}.
LRQ-Cache follows this principle but differs in two respects.
It fits low-rank factors online to each cached feature at refresh steps, so it needs no calibration and also applies to features that correspond to no single weight matrix, such as a whole-stack residual.
Unlike keys and values, which enter attention directly, a cached diffusion feature enters a reuse operator that may extrapolate it over several steps, so the reconstruction error of a finite difference is scaled by the extrapolation coefficient.
Within diffusion inference, CompactFusion compresses stepwise activation differences, but to reduce inter-device communication rather than stored state~\cite{luo2025compactfusion}.

\section{Formulation and Motivation}
\label{sec:prelim}
We first formalize residual and predictive caching (Sec.~\ref{sec:cache_formulation}), then quantify the state they keep in memory (Sec.~\ref{sec:storage_motivation}), and finally analyze the structure of this state that LRQ-Cache exploits (Sec.~\ref{sec:structural_argument}).

\subsection{Feature caching}
\label{sec:cache_formulation}
Each of the $L$ layers of a video DiT updates the hidden states of all tokens through attention and MLP modules whose outputs are added back to the hidden states.
We call the entries of a token's hidden state its channels, so that every feature cached in this process is a matrix $X\in\mathbb R^{N\times D}$ with $N$ tokens and $D$ channels.

Feature caching builds on this additive structure: it computes the network in full only at refresh (full) steps and, at the reuse steps in between, replaces part of this computation with features cached at the last refresh.
The mechanisms we consider can all be described through a module $f$ of the network whose output is added to its input hidden states $x$:
\begin{equation}
    x\leftarrow x+g_t\odot Y_t,
    \qquad
    Y_t=f_t(x),
    \label{eq:cache}
\end{equation}
where the subscript $t$ denotes dependence on the timestep and the conditioning, and $g_t$ is a gate computed from the timestep embedding, with $g_t=1$ for modules without one.
At a refresh step $s$, $Y_s$ is computed and cached; at a later reuse step $t$, the module is skipped and $Y_t$ is replaced by an estimate from its cache $\mathcal C_s$,
\begin{equation}
    \tilde Y_t=Y_s+\kappa_t\Delta_s,
    \label{eq:predict}
\end{equation}
where $\Delta_s$ is an optional finite difference and $\kappa_t$ a scalar weight determined by the mechanism.

The mechanisms differ first in what they cache. TeaCache treats the whole stack of $L$ layers as a single module: it caches the residual $Y_s=x^L_s-x^0_s$ between the output of the last layer and the input of the first and reuses it unchanged, i.e., $\kappa_t=0$~\cite{liu2025teacache}.
Predictive caches instead cache the attention and MLP modules of every layer and extrapolate their outputs over time~\cite{liu2025taylorseer,gu2026scalingcache}.
In the first-order configuration we evaluate, TaylorSeer uses the finite difference $\Delta_s=(Y_s-Y_{s'})/(s-s')$ over the previous refresh $s'$ with $\kappa_t=t-s$, while ScalingCache rescales the same difference by calibrated per-step coefficients and sets $\kappa_t$ to a sum of their cumulative products.

They also differ in when they refresh.
TaylorSeer refreshes at a fixed interval, whereas TeaCache refreshes once the accumulated change of a probe signal exceeds a threshold, and ScalingCache refreshes once the accumulated change of module outputs exceeds a threshold derived from its own history.
Throughout, we call the caches $\mathcal C_s=\{Y_s,\Delta_s\}$ of all modules the cached feature state, and the quantities kept only for refresh decisions, such as TeaCache's probe and ScalingCache's coefficients and error records, the controller state; LRQ-Cache compresses the former and leaves the latter unchanged; together they form the complete persistent cache.

\subsection{Size of the cached state}
\label{sec:storage_motivation}
To quantify this state, let $\mathcal M$ be the set of modules whose outputs are cached: TeaCache caches a single module spanning all layers, whereas predictive caches cache the attention and MLP modules of every layer.
Each module keeps one cache $\mathcal C_s$ (Sec.~\ref{sec:cache_formulation}) with $|\mathcal C|$ features, namely $Y_s$ alone for TeaCache ($|\mathcal C|=1$) and $Y_s$ together with $\Delta_s$ for first-order predictive caches ($|\mathcal C|=2$).
These caches are kept separately for each of the $G$ network evaluations per step, e.g., the conditional and unconditional branches of classifier-free guidance.
Native caches store each element of these $N\times D$ features in BF16, i.e., in two bytes.
The cached feature state therefore occupies
\begin{equation}
    S_{\rm native}=2\,|\mathcal C|\,G\,|\mathcal M|\,ND
    \label{eq:state_bytes}
\end{equation}
bytes, excluding the controller state, which we report separately (Appendix~\ref{sec:storage_analysis}).
The factors $|\mathcal C|$, $G$, $|\mathcal M|$, and $D$ are fixed by the model and the mechanism, whereas the number of tokens $N$ grows linearly with the number of frames and with the frame area.
For example, for TaylorSeer on Wan2.1, with $G=2$, $|\mathcal M|=120$ (self-attention, cross-attention, and MLP modules in each of 40 layers), $|\mathcal C|=2$, and $D=5120$, Eq.~\ref{eq:state_bytes} gives 150\,GiB at $832\times480$ with 81 frames, which matches the measured state (Fig.~\ref{fig:state_scaling} in Appendix~\ref{sec:storage_analysis}).

\subsection{Redundancy within cached features}
\label{sec:structural_argument}
\begin{figure}[t]
    \centering
    \includegraphics[width=\columnwidth]{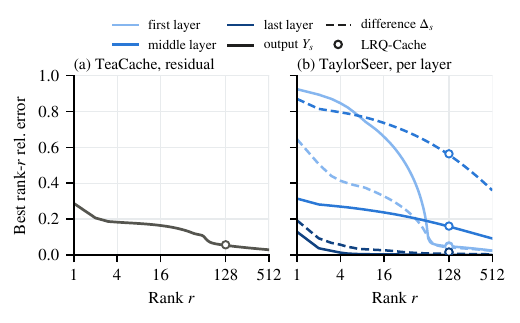}
    \caption{\textbf{How well cached features of HunyuanVideo compress.} Relative error of the best rank-$r$ approximation; lower means more compressible. (a) TeaCache residual. (b) TaylorSeer MLP features of three layers: outputs $Y_s$ (solid) and finite differences $\Delta_s$ (dashed). Lines are medians over two prompts at three refreshes; circles are LRQ-Cache (rank 128, INT8).}
    \label{fig:tea_spectrum}
\end{figure}
Since LRQ-Cache leaves the model, the caching mechanism, and the video length and resolution unchanged, the only term of Eq.~\ref{eq:state_bytes} it can reduce is the $2ND$ bytes spent on each feature, so we examine whether a single cached feature $X\in\mathbb R^{N\times D}$ is itself redundant.
To measure this redundancy, let $\mu=\frac1N X^{\mathsf T}\mathbf 1$ be its mean feature, i.e., the channel pattern shared by all tokens, and let $\rho=N\|\mu\|_2^2/\|X\|_F^2$ be the fraction of the squared Frobenius norm of $X$ that this pattern accounts for when repeated at every token.
Repeating $\mu$ at every token gives the rank-one matrix $\mathbf 1\mu^{\mathsf T}$, so a feature with large $\rho$ is close to rank one, and further components can describe how tokens deviate from it.
This suggests storing $X$ as a small set of channel patterns together with per-token coefficients: the $r$ columns of $V\in\mathbb R^{D\times r}$ are channel patterns, and each row of $A\in\mathbb R^{N\times r}$ combines them at one token, so that
\begin{equation}
    X\approx AV^{\mathsf T}.
    \label{eq:factor}
\end{equation}
The best rank-$r$ approximation $X_r$ of $X$ is therefore at least as accurate as $\mathbf 1\mu^{\mathsf T}$ for every $r\geq1$, so its relative error $\|X-X_r\|_F/\|X\|_F$ is at most $\sqrt{1-\rho}$, and higher ranks also capture the variation around the mean (Appendix~\ref{sec:shared_analysis}).

To test this argument, we measure the full singular spectrum of the TeaCache residual of HunyuanVideo at $1280\times720$ with 129 frames, for two evaluation prompts at the first, middle, and last refresh (Fig.~\ref{fig:tea_spectrum}a).
Its median $\rho$ is 0.914, and the best rank-one approximation has a median relative error of 0.29, close to the bound $\sqrt{1-\rho}$.
The error falls to 0.054 at rank 128, where the factors hold 4.3\% of the values of $X$, and the randomized INT8 factors of LRQ-Cache (Sec.~\ref{sec:method}) reach 0.057.
Predictive caches, in contrast, are less compressible in their middle layers: for TaylorSeer on HunyuanVideo, the rank-128 error is at most 0.05 in the first and last layers but 0.16 for middle-layer outputs and 0.56 for their finite differences (Fig.~\ref{fig:tea_spectrum}b), and Wan2.1 reaches 0.60 and 0.77 (Appendix~\ref{sec:structure_appendix}), so we evaluate them end to end (Sec.~\ref{sec:experiments}).

\section{LRQ-Cache}
\label{sec:method}
\begin{table*}[t]
\centering
\caption{\textbf{Native caching with and without LRQ-Cache.} Each entry averages 50 inputs per model. State is the complete persistent cache in GiB, including controller state; Dense keeps no temporal cache. VBench is the mean of five normalized dimensions; PSNR and LPIPS compare each video with Dense inference for the same input. Shaded rows are LRQ-Cache (ours): blue numbers give the reduction of the complete state relative to the same native cache, and gray numbers the VBench change in pp. Table~\ref{tab:ablation} gives fidelity to the native cache and the ablations. Ratios and differences are computed before rounding.}
\label{tab:main}
\footnotesize
\setlength{\tabcolsep}{1.45pt}
\begin{tabular}{@{}l*{3}{r@{\,}lr@{\,}lrr}@{}}
\toprule
& \multicolumn{6}{c}{HunyuanVideo}
& \multicolumn{6}{c}{Wan2.1 T2V}
& \multicolumn{6}{c}{Wan2.1 I2V} \\
\cmidrule(lr){2-7}
\cmidrule(lr){8-13}
\cmidrule(l){14-19}
Method
& \multicolumn{1}{c}{State} &
& \multicolumn{1}{c}{VBench$\uparrow$} &
& PSNR$\uparrow$ & LPIPS$\downarrow$
& \multicolumn{1}{c}{State} &
& \multicolumn{1}{c}{VBench$\uparrow$} &
& PSNR$\uparrow$ & LPIPS$\downarrow$
& \multicolumn{1}{c}{State} &
& \multicolumn{1}{c}{VBench$\uparrow$} &
& PSNR$\uparrow$ & LPIPS$\downarrow$ \\
\midrule

Dense
& -- & & 84.71\% & & -- & --
& -- & & 84.82\% & & -- & --
& -- & & 84.42\% & & -- & -- \\
\midrule

TeaCache
& 2.04 & & 84.50\% & & 22.78 & 0.2217
& 0.63 & & 84.70\% & & 27.77 & 0.0949
& 1.44 & & 84.39\% & & 30.96 & 0.0623 \\

\rowcolor{oursbg}
\quad + LRQ-Cache
& \textbf{1.37}
& \textcolor{cvprblue}{\scriptsize 1.5$\times$}
& 84.26\%
& \textcolor{deltagray}{\scriptsize($-$0.24)}
& 22.81 & 0.2281
& \textbf{0.01}
& \textcolor{cvprblue}{\scriptsize 64$\times$}
& 84.65\%
& \textcolor{deltagray}{\scriptsize($-$0.05)}
& 26.44 & 0.1161
& \textbf{0.02}
& \textcolor{cvprblue}{\scriptsize 70$\times$}
& 84.43\%
& \textcolor{deltagray}{\scriptsize(+0.05)}
& 29.97 & 0.0707 \\
\midrule

TaylorSeer
& 109 & & 80.10\% & & 15.33 & 0.5508
& 150 & & 79.17\% & & 12.36 & 0.5918
& 346 & & 81.70\% & & 17.08 & 0.4137 \\

\rowcolor{oursbg}
\quad + LRQ-Cache
& \textbf{2.59}
& \textcolor{cvprblue}{\scriptsize 42$\times$}
& 79.02\%
& \textcolor{deltagray}{\scriptsize($-$1.08)}
& 15.38 & 0.5452
& \textbf{2.30}
& \textcolor{cvprblue}{\scriptsize 65$\times$}
& 76.04\%
& \textcolor{deltagray}{\scriptsize($-$3.12)}
& 12.24 & 0.6160
& \textbf{4.91}
& \textcolor{cvprblue}{\scriptsize 71$\times$}
& 78.18\%
& \textcolor{deltagray}{\scriptsize($-$3.53)}
& 16.39 & 0.5669 \\
\midrule

ScalingCache
& 109 & & 84.65\% & & 34.41 & 0.0596
& 150 & & 84.62\% & & 27.37 & 0.1128
& 346 & & 84.41\% & & 29.41 & 0.0813 \\

\rowcolor{oursbg}
\quad + LRQ-Cache
& \textbf{2.51}
& \textcolor{cvprblue}{\scriptsize 43$\times$}
& 84.54\%
& \textcolor{deltagray}{\scriptsize($-$0.11)}
& 34.76 & 0.0606
& \textbf{2.30}
& \textcolor{cvprblue}{\scriptsize 65$\times$}
& 81.84\%
& \textcolor{deltagray}{\scriptsize($-$2.78)}
& 24.34 & 0.2898
& \textbf{4.91}
& \textcolor{cvprblue}{\scriptsize 71$\times$}
& 82.97\%
& \textcolor{deltagray}{\scriptsize($-$1.44)}
& 25.27 & 0.2877 \\

\bottomrule
\end{tabular}
\end{table*}
Building on this redundancy, LRQ-Cache changes only how the cached state is stored (Fig.~\ref{fig:overview}(b)).
It compresses every video-token feature that a mechanism keeps in its caches $\mathcal C_s$ (features of at most 512 tokens stay native; Appendix~\ref{sec:setup_details}), namely the residual $Y_s$ of TeaCache and the features $Y_s$ and finite differences $\Delta_s$ of predictive caches.
Concretely, at each refresh step, it factorizes each such feature into low-rank factors (Sec.~\ref{sec:lowrank}) and quantizes the factors to INT8 (Sec.~\ref{sec:factor_quant}); at each reuse step, it reconstructs the feature for the unchanged estimate of Eq.~\ref{eq:predict} (Sec.~\ref{sec:reuse}).
All other components, i.e., the model, the refresh rules, the predictors, and the controller state, are left unchanged; adaptive rules can nevertheless refresh at different steps because reconstruction changes the hidden states they observe (Sec.~\ref{sec:adaptive}).

\subsection{Low-rank factorization}
\label{sec:lowrank}
At each refresh step, LRQ-Cache keeps the leading $r$ singular components of every cached feature $X$ and stores them as two factors,
\begin{equation}
    X\approx U_r\Sigma_rV_r^{\mathsf T}=AV^{\mathsf T},
    \qquad
    A=U_r\Sigma_r,\quad V=V_r,
    \label{eq:svd}
\end{equation}
where $\Sigma_r=\operatorname{diag}(\sigma_1,\dots,\sigma_r)$ holds the $r$ largest singular values of $X$, and the columns of $U_r\in\mathbb R^{N\times r}$ and $V_r\in\mathbb R^{D\times r}$ are the corresponding singular vectors.
By construction, the columns of $V$ are orthonormal, and the $i$-th column of $A$ has norm $\sigma_i$.
In practice, we compute the factors with randomized low-rank factorization~\cite{halko2011finding}, which avoids a full SVD.
We factorize $X$ itself rather than its centered remainder: by Sec.~\ref{sec:structural_argument}, the best rank-$r$ approximation of $X$ is already at least as accurate as the shared pattern $\mathbf 1\mu^{\mathsf T}$, so $\mu$ need not be stored separately.

We factorize the cached features online rather than the network weights offline as in KV-cache compression~\cite{chang2025palu}: this leaves every full evaluation unchanged, needs no calibration, and fits the factors to spectra that differ across layers and refreshes, and the whole-stack residual of TeaCache has no single weight matrix to factorize.
Encoding thus runs at refresh steps and decoding at reuse steps (Sec.~\ref{sec:codec_cost}).
The rank $r$ is the only parameter of LRQ-Cache: the state grows linearly with $r$, so $r$ sets the trade-off between memory and fidelity; we use $r{=}128$ unless noted.

\subsection{Quantizing the factors}
\label{sec:factor_quant}
The factors already shrink each feature, but stored in BF16 they would still occupy $2r(N+D)$ bytes; 8-bit codes halve this, plus an offset and a scale for every 128 entries.
Because the column norms $\sigma_i$ of $A$ fall across components, a range shared across components would be set by the leading ones and would quantize the trailing ones coarsely.
We therefore quantize each column of $A$ separately, and $V$ in the same layout, in groups of 128 consecutive entries.

Specifically, a group with minimum $\ell$ and maximum $u$ stores an FP32 offset $\ell$ and an FP32 scale $s=(u-\ell)/255$, and each entry $x$ is encoded to a code $q$ and decoded to $\widehat x$ as
\begin{equation}
    q=\operatorname{round}\big((x-\ell)/s\big),
    \qquad
    \widehat x=\ell+sq,
    \qquad 0\leq q\leq255.
    \label{eq:quantizer}
\end{equation}
A packet thus holds the codes, offsets, and scales of $A$ and $V$ and occupies
\begin{equation}
    B_{\rm LRQ}=r(N+D)+B_{\rm scales}
    \label{eq:packet_bytes}
\end{equation}
bytes, where $B_{\rm scales}=8r(\lceil N/128\rceil+\lceil D/128\rceil)$ counts the offsets and scales.
Compared with $2ND$ bytes for the native feature, a packet grows by $r$ bytes per token instead of $2D$ (Fig.~\ref{fig:state_scaling} in Appendix~\ref{sec:storage_analysis}).
This saving comes at the cost of a rounding error on top of the truncation error of Sec.~\ref{sec:lowrank}; Appendix~\ref{sec:factor_analysis} bounds the two separately, and Sec.~\ref{sec:ablation} measures them.

\begin{figure*}[t]
    \centering
    \includegraphics[width=\textwidth]{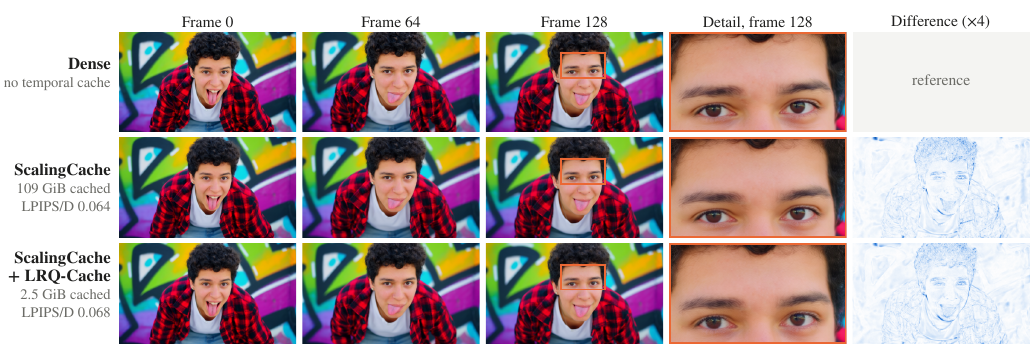}
    \caption{\textbf{Paired frames of HunyuanVideo with ScalingCache.}
    Rows show Dense inference, ScalingCache with its native cache, and ScalingCache with LRQ-Cache, for one input chosen for coherent motion in the native video.
    The last column is the mean absolute RGB difference to Dense at frame 128, magnified four times, and LPIPS/D is the whole-video LPIPS to Dense.}
    \label{fig:qualitative_primary}
\end{figure*}

\subsection{Reusing the compressed cache}
\label{sec:reuse}
At each reuse step, LRQ-Cache decodes the factors with Eq.~\ref{eq:quantizer}, multiplies them in floating point, and casts the product to the dtype of the native cache; only the packets persist between steps (Algorithm~\ref{alg:lrq}, Appendix~\ref{sec:lifecycle}).
The reconstructed $\hat Y_s$, and $\hat\Delta_s$ if present, then replace the native tensors in Eq.~\ref{eq:predict}.
One subtlety arises in predictive caches: the finite difference at a refresh is computed against the reconstructed previous output $\hat Y_{s'}$, since the native one is no longer stored, so its reconstruction error propagates into $\hat\Delta_s$; unhatted tensors denote the native reference.

Because the reconstructed tensors enter the reuse operator, reconstruction error reaches the reused output differently depending on $\kappa_t$.
Substituting the reconstructed tensors into Eq.~\ref{eq:predict} gives
\begin{equation}
    \big\|\hat Y_s+\kappa_t\hat\Delta_s-\tilde Y_t\big\|_F
    \leq\|\hat Y_s-Y_s\|_F+|\kappa_t|\,\|\hat\Delta_s-\Delta_s\|_F.
    \label{eq:reuse_gain}
\end{equation}
For TeaCache, $\kappa_t=0$ and the reconstruction error of the residual passes to the output unchanged.
For predictive caches, by contrast, the error of the finite difference is amplified by $|\kappa_t|$, which grows with the distance from the last refresh.
Finite differences can also have heavier spectral tails than outputs (Sec.~\ref{sec:structural_argument}), and hence larger truncation errors at the same rank.
Beyond this bound, reconstruction also changes later hidden states and, for adaptive mechanisms, refresh decisions, so all main comparisons use end-to-end generation.

\section{Experiments}
\label{sec:experiments}

\subsection{Setup}
\label{sec:protocol}
\paragraph{Models and inputs.}
We evaluate three state-of-the-art open-source video DiTs: HunyuanVideo~\cite{kong2024hunyuanvideo} at 720p, Wan2.1-T2V-14B~\cite{wan2025wan} at 480p, and Wan2.1-I2V-14B~\cite{wan2025wan} at 720p, each with its default number of frames.
Text-to-video inputs are 50 prompts sampled at random from the widely used VBench benchmark~\cite{huang2024vbench}, and image-to-video inputs are 50 image--text pairs sampled from VBench-I2V~\cite{huang2026vbenchpp}.

\paragraph{Compared methods.} We apply LRQ-Cache to three widely adopted caching mechanisms, the residual cache TeaCache~\cite{liu2025teacache} and the predictive caches TaylorSeer~\cite{liu2025taylorseer} and ScalingCache~\cite{gu2026scalingcache}, and compare each with its native cache and with Dense inference.
To separate the two components of LRQ-Cache, the ablation adds INT4, which quantizes the dense tensor without factorization, and SVD128, which keeps the rank-128 factors in BF16 without quantization.
Runs on the same input share the same configuration and differ only in how the cached features are stored.

\paragraph{Metrics.}
We measure fidelity against both Dense inference and the native cache: fidelity to the native cache isolates the error added by storage, whereas fidelity to Dense also contains the error of caching itself.
Fidelity is reported with PSNR, SSIM~\cite{wang2004ssim}, and LPIPS~\cite{zhang2018lpips} with VGG features (AlexNet features in Appendix~\ref{sec:lpips_alex}).
Overall quality is the mean of five video-quality dimensions computed with the official VBench implementation~\cite{huang2024vbench}, namely subject consistency, background consistency, motion smoothness, aesthetic quality, and imaging quality, reported in percent; changes are absolute differences in percentage points (pp), not relative changes; Appendix~\ref{sec:setup_details} describes each dimension and why the semantic dimensions are omitted.
Storage counts the complete persistent cache, including scales and controller state, relative to the same native cache.

\paragraph{Implementation details.}
All runs use seed 0 and the default sampler and number of steps of each model, and each mechanism is ported from its released code with its released settings.
INT4 applies the quantizer of Eq.~\ref{eq:quantizer} at 4 bits in groups of 128 tokens within each channel, and Appendix~\ref{sec:setup_details} provides the remaining settings.
LRQ-Cache uses rank 128 for every model and mechanism.
At this rank, each token keeps 128 bytes instead of $2D$, which makes the main results a test of an aggressive storage budget.
A later rank sweep, reported in Appendix~\ref{sec:equal_bytes}, places this rank at the knee of the storage--quality curve: it preserves the quality of the residual cache, whereas lower ranks degrade it sharply.
The predictive caches of Wan2.1 lose more at this rank and benefit from higher ranks, as the ablation in Sec.~\ref{sec:ablation} shows.

\begin{table*}[t]
\centering
\caption{\textbf{Quantization-only baseline and factor-precision ablation.} Each representation is compared with the same native cache over 50 inputs. State is the complete persistent cache as a percentage of native, and $\Delta$VBench the VBench change in pp. Bold marks the smallest state for each mechanism and model; ratios and differences are computed before rounding.}
\label{tab:ablation}
\footnotesize
\newcommand{\gain}[1]{\,\rlap{\textcolor{cvprblue}{\scriptsize #1$\times$}}\hphantom{\scriptsize 1.1$\times$}}
\newcommand{\nogain}{\,\hphantom{\scriptsize 1.1$\times$}}
\setlength{\tabcolsep}{3.4pt}
\begin{tabular}{@{}llrrrrrrrrr@{}}
\toprule
 & & \multicolumn{3}{c}{HunyuanVideo} & \multicolumn{3}{c}{Wan2.1 T2V} & \multicolumn{3}{c@{}}{Wan2.1 I2V} \\
\cmidrule(lr){3-5}\cmidrule(lr){6-8}\cmidrule(l){9-11}
Mechanism & Cache storage & State & LPIPS/N$\downarrow$ & $\Delta$VBench$\uparrow$ & State & LPIPS/N$\downarrow$ & $\Delta$VBench$\uparrow$ & State & LPIPS/N$\downarrow$ & $\Delta$VBench$\uparrow$ \\
\midrule
TeaCache & + INT4 & 76.04\% & 0.0175 & +0.00 & 28.15\% & 0.0409 & $-$0.01 & 28.14\% & 0.0507 & +0.01 \\
TeaCache & + SVD128 (BF16) & 68.09\% & 0.0490 & $-$0.22 & 2.93\% & 0.0780 & $-$0.07 & 2.68\% & 0.0621 & +0.05 \\
\rowcolor{oursbg}TeaCache & + LRQ-Cache (ours) & \textbf{67.42\%} & 0.0493 & $-$0.24 & \textbf{1.57\%} & 0.0784 & $-$0.05 & \textbf{1.43\%} & 0.0616 & +0.05 \\
\midrule
TaylorSeer & + INT4 & 28.21\% & 0.0884 & $-$0.14 & 28.13\% & 0.3761 & $-$4.54 & 28.13\% & 0.5399 & $-$4.37 \\
TaylorSeer & + SVD128 (BF16) & 4.38\% & 0.0917 & $-$1.08 & 2.89\% & 0.3444 & $-$3.07 & 2.67\% & 0.4867 & $-$3.48 \\
\rowcolor{oursbg}TaylorSeer & + LRQ-Cache (ours) & \textbf{2.38\%} & 0.0911 & $-$1.08 & \textbf{1.54\%} & 0.3450 & $-$3.12 & \textbf{1.42\%} & 0.4883 & $-$3.53 \\
\midrule
ScalingCache & + INT4 & 28.13\% & 0.0158 & +0.02 & 28.13\% & 0.1480 & $-$1.55 & 28.13\% & 0.1528 & $-$0.71 \\
ScalingCache & + SVD128 (BF16) & 4.33\% & 0.0331 & $-$0.10 & 2.89\% & 0.2790 & $-$2.82 & 2.67\% & 0.2839 & $-$1.46 \\
\rowcolor{oursbg}ScalingCache & + LRQ-Cache (ours) & \textbf{2.30\%} & 0.0345 & $-$0.11 & \textbf{1.54\%} & 0.2794 & $-$2.78 & \textbf{1.42\%} & 0.2838 & $-$1.44 \\
\bottomrule
\end{tabular}
\end{table*}

\subsection{Comparison with native caching}
\label{sec:main_results}
As Table~\ref{tab:main} shows, LRQ-Cache shrinks the cached state to under 2.5\% of its native size, since each token keeps one byte per low-rank component instead of two bytes per channel.
The only exception is HunyuanVideo TeaCache, whose uncompressed FP32 probe, twice the size of the residual, limits the reduction to a third.
For the predictive caches, the state falls from more than 100\,GiB to under 5\,GiB per video, small enough to stay on one GPU beside the model weights.

For TeaCache and for the predictive caches of HunyuanVideo, quality barely changes.
VBench stays within 0.25\,pp of the native cache except for a drop of about 1\,pp on HunyuanVideo TaylorSeer, and only TeaCache on Wan2.1 drifts noticeably farther from Dense in LPIPS.
Figure~\ref{fig:qualitative_primary} and Fig.~\ref{fig:qualitative_wan} in the appendix show paired frames from two of these settings.
Little reconstruction error reaches the output here, because TeaCache reuses its residual without extrapolation and the predictive caches of HunyuanVideo keep much smaller middle-layer errors than those of Wan2.1.
In these settings, the memory reduction thus comes almost for free.

The predictive caches of Wan2.1 lose more, up to 3.5\,pp of VBench.
As Section~\ref{sec:structural_argument} shows, their middle-layer features keep large errors at rank 128, and extrapolation amplifies the error of each finite difference with the distance from the last refresh, so TaylorSeer, which refreshes only every fifth step, loses the most.
Changed refresh decisions do not explain the loss: TaylorSeer refreshes at fixed steps, and ScalingCache recovers most of its loss at a higher rank although its refresh steps still differ from native.
As Section~\ref{sec:ablation} shows, the loss is set by the rank rather than by INT8 rounding, and a higher rank recovers most of it.

\subsection{Ablation}
\label{sec:ablation}
Table~\ref{tab:ablation} isolates the two steps of LRQ-Cache with INT4, which quantizes the dense tensor without factorizing it, and SVD128, which factorizes it but keeps the factors in BF16, each compared with the native cache of the same mechanism.

\paragraph{Quantization alone.}
Quantization keeps the number of stored values, so a dense tensor with $b$-bit codes keeps at least $b/16$ of its BF16 size, and even 1-bit codes would keep over twice the feature bytes of LRQ-Cache.
We use 4 bits, which keeps more than a quarter of the state, an order of magnitude more than LRQ-Cache, whereas 8 bits would keep more than half.
With this larger budget, INT4 stays closer to the native cache except on the Wan2.1 TaylorSeer caches, where it also loses more VBench.
Quantization alone suffices when the compressible part of a cache is small, as on HunyuanVideo TeaCache, whose uncompressed probe makes both complete states similar in size.
LRQ-Cache reduces the number of stored values instead, which takes it below this floor: on Wan2.1 T2V TeaCache, rank 256 stays closer to the native cache than 2-bit codes at a fifth of their state, as the sweep in Appendix~\ref{sec:equal_bytes} shows.

\paragraph{Factor precision.}
INT8 factors nearly halve the state of SVD128, except on HunyuanVideo TeaCache, where the probe dominates.
VBench nevertheless stays within 0.1\,pp of SVD128 in every setting, although LPIPS shifts in both directions on individual videos.
We stop at 8 bits because, in the spectra of Section~\ref{sec:structural_argument}, 8-bit rounding adds almost nothing to the truncation error, and fewer bits would save only about another 1\% of the native state.
Since the Wan2.1 predictive caches already lose quality with BF16 factors, the loss of LRQ-Cache comes from the rank-128 approximation rather than from INT8 rounding.

\paragraph{Rank.}
Raising the rank trades state for fidelity.
Table~\ref{tab:rank} reruns both Wan2.1 T2V predictive caches at ranks fixed in advance: at rank 512, which keeps about 6\% of the native state, TaylorSeer recovers more than half of its VBench loss and ScalingCache about three quarters, and both come closer to the native cache than INT4 at under a quarter of its state.
The returns diminish, however: for both caches, the second doubling of the rank recovers about half as much VBench as the first, while encoding time more than triples from rank 128 to 512.
Since the error and its amplification differ across cache sites, allocating rank per site may give a better trade-off.

\begin{table}[t]
\centering
\caption{\textbf{Rank on Wan2.1 T2V predictive caches.} $r$: rank. State: complete persistent cache, \% of native; LPIPS and $\Delta$VBench (pp) to the same native cache; 50 prompts each (Appendix~\ref{sec:equal_bytes}).}
\label{tab:rank}
\footnotesize
\setlength{\tabcolsep}{1.9pt}
\begin{tabular}{@{}lrrrrr@{}}
\toprule
 & & \multicolumn{2}{c}{TaylorSeer} & \multicolumn{2}{c@{}}{ScalingCache} \\
\cmidrule(lr){3-4}\cmidrule(l){5-6}
Representation & State & LPIPS$\downarrow$ & $\Delta$VBench & LPIPS$\downarrow$ & $\Delta$VBench \\
\midrule
+ INT4 & 28.13\% & 0.3761 & $-$4.54 & 0.1480 & $-$1.55 \\
+ LRQ-Cache ($r{=}128$) & 1.54\% & 0.3450 & $-$3.12 & 0.2794 & $-$2.78 \\
+ LRQ-Cache ($r{=}256$) & 3.07\% & 0.2970 & $-$1.85 & 0.1835 & $-$1.34 \\
+ LRQ-Cache ($r{=}512$) & 6.14\% & \textbf{0.2581} & \textbf{$-$1.19} & \textbf{0.1125} & \textbf{$-$0.64} \\
\bottomrule
\end{tabular}
\end{table}

\subsection{Analysis}
\label{sec:analysis}
\paragraph{Adaptive execution.}
\label{sec:adaptive}
Under LRQ-Cache, TeaCache and TaylorSeer keep their native full steps on HunyuanVideo and Wan2.1 T2V and their number on Wan2.1 I2V, since they refresh at a fixed interval or from an uncompressed probe.
ScalingCache instead computes its criterion from the decoded features and refreshes at different steps on almost every input.
The comparison with the native cache thus isolates storage for TeaCache and TaylorSeer, and Section~\ref{sec:main_results} shows that the shifted schedule of ScalingCache does not drive its quality loss.

\paragraph{Single-GPU time and memory.}
\label{sec:single_gpu}
As Table~\ref{tab:single_gpu} shows, neither the native predictive states nor their INT4 versions fit on one 80\,GB H100 beside the model weights.
Offloaded to host memory, the native caches spend most of their time on the synchronous copies of our baseline, which makes them slower than Dense in most settings.
LRQ-Cache keeps its state on the GPU and runs 2.7--5.9$\times$ faster, restoring the speedup of predictive caching.
Pinned, overlapped copies would narrow this gap, but only LRQ-Cache removes the need for more than 100\,GiB of host memory per video.

\begin{table}[t]
\centering
\caption{\textbf{Single-GPU generation time and memory.} One 80\,GB H100 with the model weights resident. Times are seconds per video, medians over three inputs and two repetitions, one for native on Wan2.1 I2V; $\times$Nat.\ is the speedup of LRQ-Cache over native. Native states stay in host memory because they, like INT4, run out of memory on the GPU. GPU is the peak allocated memory during denoising for Dense / LRQ-Cache and Host the peak host memory for native / LRQ-Cache, both in GiB. Quality is reported in Table~\ref{tab:main}.}
\label{tab:single_gpu}
\small
\setlength{\tabcolsep}{2.0pt}
\begin{tabular}{@{}lrrrrcc@{}}
\toprule
Mechanism & Dense & Native & LRQ & $\times$Nat. & GPU & Host \\
\midrule
\multicolumn{7}{@{}l}{\textit{HunyuanVideo}} \\
\quad TaylorSeer & 1844 & 1573 & \textbf{511} & 3.1$\times$ & 57.7/60.3 & 113.0/6.2 \\
\quad ScalingCache & 1841 & 2775 & \textbf{1026} & 2.7$\times$ & 57.8/60.3 & 113.1/6.7 \\
\midrule
\multicolumn{7}{@{}l}{\textit{Wan2.1 T2V}} \\
\quad TaylorSeer & 498 & 1678 & \textbf{309} & 5.4$\times$ & 41.4/43.4 & 152.5/2.6 \\
\quad ScalingCache & 498 & 2456 & \textbf{564} & 4.4$\times$ & 41.4/43.4 & 152.5/2.7 \\
\midrule
\multicolumn{7}{@{}l}{\textit{Wan2.1 I2V}} \\
\quad TaylorSeer & 1569 & 3871 & \textbf{661} & 5.9$\times$ & 52.4/56.6 & 351.0/5.3 \\
\quad ScalingCache & 1569 & 6148 & \textbf{1418} & 4.6$\times$ & 52.4/56.6 & 350.9/5.3 \\
\bottomrule
\end{tabular}
\end{table}

\paragraph{Codec cost.}
\label{sec:codec_cost}
Encoding is the main overhead of LRQ-Cache, since each refresh factorizes every cached feature.
On Wan2.1 T2V, it costs TeaCache only seconds per video, whereas the predictive caches encode hundreds of features per refresh and spend over half of their time in the codec.
Since ScalingCache refreshes about twice as often as TaylorSeer, this overhead makes it slower than Dense on Wan2.1 T2V; in every other setting, the skipped steps save more time than the codec costs.

\section{Limitations and Future Work}
\label{sec:discussion}
\paragraph{Costs of compression.}
LRQ-Cache trades fidelity and encoding time for memory.
The fidelity cost falls on predictive caches, whose finite differences compress worse than outputs and are amplified by the reuse operator, so VBench drops by up to 3.5\,pp on Wan2.1, whereas TeaCache stays within 0.25\,pp.
Rank sets this trade-off: as Table~\ref{tab:rank} shows, rank 512 recovers most of the VBench loss on Wan2.1 T2V at about 6\% of the native state, but at more than three times the encoding time.

\paragraph{Scope of the evaluation.}
The spectra measure how well the cached tensors of each layer are approximated, not video quality, so they do not show how the error of each layer affects the video.
Single-GPU timings cover six settings and three inputs, and rank sweeps cover only Wan2.1 T2V.

\paragraph{Future work.}
LRQ-Cache uses one rank for all layers, although the spectra show a small error in the first and last layers and a large one in the middle (at rank 128, at most 0.05 versus up to 0.56; Sec.~\ref{sec:structural_argument}).
A layer-wise rank that follows this error, lower at the ends and higher in the middle, may give a better trade-off.

\section{Conclusion}
Feature caching accelerates video DiTs, but the features it keeps for reuse can outgrow a single GPU.
Because these features are redundant across tokens, LRQ-Cache stores them as INT8-quantized low-rank factors, without training and without changing the model or the caching mechanism.
On HunyuanVideo and Wan2.1, this shrinks the state of predictive caches by 42--71$\times$, enough to run them on one GPU 2.7--5.9$\times$ faster than with CPU offloading.
Quality stays close to the native cache except for the predictive caches of Wan2.1, whose loss shrinks as the rank grows.
Section~\ref{sec:discussion} discusses the remaining limitations.

\section*{LLM Usage Statement}
We used the LLM-based assistants OpenAI Codex and Anthropic Claude Code.
For code and experiments, they wrote parts of the method implementation and of the experiment scripts, and helped monitor experiment jobs.
For text and figures, they revised the presentation of the paper, edited the related-work descriptions, checked the bibliography entries, and refined the styling of figures and tables.
They also gave reviewer-style feedback on drafts of the manuscript.
All reported numbers come from logged runs; no LLM produced, estimated, or annotated any data.
The authors reviewed all AI-assisted content and take full responsibility for the submission.
{\small
\bibliographystyle{ieeenat_fullname}
\bibliography{main}
}

\clearpage
\appendix
\twocolumn[\centering\Large\bfseries Appendix\par\vspace{1.0em}]
\section{Cache lifecycle}
\label{sec:lifecycle}
Algorithm~\ref{alg:lrq} follows one cached module through one network evaluation per step, as our runs implement it; for TeaCache, $f$ is the whole stack, $Y_s$ is its residual, and no difference is stored.
Only $Y_s$ is encoded at the first refresh, so TaylorSeer reuses $\hat Y_s$ alone until its second refresh and ScalingCache forecasts only after its second; from then on each refresh measures the difference against the current uncompressed output and the decoded previous output, and the new packets replace the old ones.
Compression therefore reaches the reuse steps and the later differences, while the refresh step itself and the controller state stay uncompressed; TeaCache compares its timestep-modulation probe, and ScalingCache forms its relative $\ell_1$ error records from decoded tensors.

\begin{algorithm*}[!t]
\caption{LRQ-Cache for one cached module and one network evaluation per step. $\mathcal E$ encodes a feature into a packet (Eqs.~\ref{eq:svd} and~\ref{eq:quantizer}); $\mathcal D$ decodes it.}
\label{alg:lrq}
\small
\begin{algorithmic}[1]
\State $P_Y\gets\varnothing$, $P_\Delta\gets\varnothing$
\For{each step $t$}
  \State $\mathit{full}\gets$ refresh rule of the mechanism on its uncompressed controller state
  \If{$\mathit{full}$} \Comment{refresh step $s=t$}
    \State $Y_s\gets f_s(x)$, used unchanged in Eq.~\ref{eq:cache}
    \If{the cache is predictive and $P_Y\neq\varnothing$} \Comment{previous refresh $s'$}
      \State $\hat Y_{s'}\gets\mathcal D(P_Y)$
      \State ScalingCache only: record the error of $\hat Y_{s'}$ against $Y_s$
      \State $P_\Delta\gets\mathcal E\big(c_{s',s}\,(Y_s-\hat Y_{s'})\big)$ \Comment{TaylorSeer: $c_{s',s}=1/(s-s')$}
    \EndIf
    \State $P_Y\gets\mathcal E(Y_s)$ \Comment{old packets are dropped}
  \Else \Comment{reuse step}
    \State $\hat Y_s\gets\mathcal D(P_Y)$; \ $\hat\Delta_s\gets\mathcal D(P_\Delta)$ if $P_\Delta\neq\varnothing$, else $0$
    \State $\tilde Y_t\gets\hat Y_s+\kappa_t\hat\Delta_s$ in place of $f_t(x)$ (Eq.~\ref{eq:predict})
    \State ScalingCache only: record the error of $\hat Y_s$ against $\tilde Y_t$
  \EndIf
\EndFor
\end{algorithmic}
\end{algorithm*}

\section{Experimental details}
\label{sec:setup_details}
\paragraph{Inputs and sampling.}
The text-to-video prompts are sampled from the 946 GPT-4o-enhanced prompts of VBench~\cite{huang2024vbench}, and the image-to-video inputs from the 355 image--text pairs of VBench-I2V~\cite{huang2026vbenchpp}, with the official 16:9 crops.
HunyuanVideo uses reverse flow Euler sampling with shift 7 and embedded guidance 6, at 24 fps; both Wan2.1 models use UniPC with CFG 5 and separate conditional and unconditional branches, with shift 3 for T2V and 5 for I2V, at 16 fps.

\paragraph{Caching mechanisms.}
TaylorSeer predicts to first order and takes a full step every fifth step, while TeaCache and ScalingCache use their released coefficients for each model; on Wan2.1 I2V, where ScalingCache releases none, we use its Wan2.1-T2V-14B coefficients.
Native TeaCache, TaylorSeer, and ScalingCache compute 22, 10, and 22.34 of 50 steps in full on HunyuanVideo and 27, 10, and 18.52 on Wan2.1 T2V, and 30, 8, and 18.88 of 40 on Wan2.1 I2V.
For ScalingCache, the mean number of full steps for SVD128, INT4, and LRQ-Cache is 20.60, 19.84, and 20.58 on Wan2.1 T2V, 21.84, 22.00, and 21.82 on HunyuanVideo, and 20.22, 19.92, and 20.20 on Wan2.1 I2V.

\paragraph{Storage representations.}
LRQ-Cache computes the randomized factorization in FP32 with oversampling 16 and two power iterations and then quantizes the FP32 factors, drawing the random projections from a separate generator so that the sampling noise of each video is unchanged.
At $r=128$ the factors keep 128 bytes per token instead of $2D$, i.e., 1.25\% of a dense Wan2.1 feature and 2.1\% of a HunyuanVideo one.
INT4 instead stores an FP32 offset and scale for every group of 128 tokens within a channel and so spends 4.5 bits per value, 28.1\% of BF16.

\paragraph{Metrics.}
PSNR is the mean over frames, SSIM uses an $11\times11$ box window, and LPIPS~\cite{zhang2018lpips} uses version 0.1 with VGG features on frames scaled to $[-1,1]$ (AlexNet features in Appendix~\ref{sec:lpips_alex}); each metric is averaged over the frames of a video and then over the inputs.
VBench scores come from the official implementation~\cite{huang2024vbench} in its custom-input mode, applied to our generated videos, for five video-quality dimensions: subject consistency, background consistency, motion smoothness, aesthetic quality, and imaging quality.
Each dimension is VBench's aggregate score in $[0,1]$ over the 50 inputs of a model, reported in percent, and our VBench score weights the five dimensions equally.
The semantic dimensions of VBench measure prompt adherence and require their own prompt suites, so we leave them out; with the official weighting omitted as well, our score is not comparable with the total score of the VBench leaderboard.

\paragraph{Storage accounting.}
Complete-cache bytes count persistent CPU and GPU tensors, scales, controller probes, and histories, with aliased storage counted once; model weights, transient reconstructions, allocator overhead, and non-tensor descriptors (at most 74,880 bytes per Wan2.1 T2V video) are excluded.
Ratios take the Native state of the same model and mechanism as the denominator, and Dense has no temporal-cache ratio.
Caches reside on GPU for HunyuanVideo TeaCache and on CPU for both Wan2.1 models and HunyuanVideo ScalingCache; HunyuanVideo TaylorSeer keeps its native cache on CPU, since its 109\,GiB exceed the memory of one 80\,GB GPU, and its compressed caches on GPU.
On Wan2.1 I2V the 346\,GiB native states of 25 TaylorSeer and 41 ScalingCache videos were held on three additional GPUs instead of host memory, a placement that reproduced the host-stored videos bit for bit in two validation runs.
The savings are therefore not proportional reductions in GPU memory.
In HunyuanVideo TaylorSeer, text-stream outputs of at most 512 tokens stay uncompressed and are counted.

\paragraph{Single-GPU runs.}
The native baseline of Table~\ref{tab:single_gpu} copies the state synchronously into newly allocated pageable tensors whenever it is written or read.
HunyuanVideo and Wan2.1 T2V were timed on one dedicated host with at most two concurrent jobs per CPU socket, and Wan2.1 I2V on a second H100 host with each native configuration run once per input; repetitions produce identical frames, and the peak GPU memory of the native runs stays within 0.7\,GiB of Dense.
These runs use a separate harness that imports the cache adapters and codecs of the quality runs but keeps all weights on the GPU.
They reproduce the full-step schedules and cache bytes of the quality runs, and their frames match those runs bit for bit for HunyuanVideo ScalingCache but differ on Wan2.1 for every configuration, including Dense, so part of the difference arises without caching; Table~\ref{tab:single_gpu} therefore reports time and memory only, and quality comes from Tables~\ref{tab:main} and~\ref{tab:ablation}.

\section{LPIPS with AlexNet features}
\label{sec:lpips_alex}
\begin{table*}[!t]
\centering
\caption{\textbf{LPIPS with AlexNet features.} Same 50 inputs per model and same pairs as Table~\ref{tab:all_representations}, with inputs scaled to $[-1,1]$. /D compares with Dense inference and /N with the same native cache; the shaded column is LRQ-Cache (ours). Bold marks the lowest LPIPS/N in each row.}
\label{tab:lpips_alex}
\small
\setlength{\tabcolsep}{4pt}
\begin{tabular}{@{}llrrrrr@{}}
\toprule
 & & \multicolumn{2}{c}{LPIPS/D$\downarrow$} & \multicolumn{3}{c}{LPIPS/N$\downarrow$} \\
\cmidrule(lr){3-4}\cmidrule(l){5-7}
Model & Mechanism & Native & LRQ-Cache & INT4 & SVD128 & LRQ-Cache \\
\midrule
HunyuanVideo & TeaCache & 0.1674 & 0.1731 & \textbf{0.0077} & 0.0259 & \cellcolor{oursbg}0.0260 \\
 & TaylorSeer & 0.5142 & 0.5234 & \textbf{0.0284} & 0.0451 & \cellcolor{oursbg}0.0444 \\
 & ScalingCache & 0.0311 & 0.0337 & \textbf{0.0049} & 0.0137 & \cellcolor{oursbg}0.0146 \\
\midrule
Wan2.1 T2V & TeaCache & 0.0587 & 0.0712 & \textbf{0.0210} & 0.0418 & \cellcolor{oursbg}0.0421 \\
 & TaylorSeer & 0.5842 & 0.5899 & 0.2556 & \textbf{0.2210} & \cellcolor{oursbg}0.2212 \\
 & ScalingCache & 0.0657 & 0.1660 & \textbf{0.0873} & 0.1620 & \cellcolor{oursbg}0.1622 \\
\midrule
Wan2.1 I2V & TeaCache & 0.0337 & 0.0388 & \textbf{0.0254} & 0.0324 & \cellcolor{oursbg}0.0321 \\
 & TaylorSeer & 0.3426 & 0.4475 & 0.4420 & \textbf{0.3480} & \cellcolor{oursbg}0.3493 \\
 & ScalingCache & 0.0456 & 0.1671 & \textbf{0.0919} & 0.1664 & \cellcolor{oursbg}0.1651 \\
\bottomrule
\end{tabular}
\end{table*}
The main text reports LPIPS with VGG features; Table~\ref{tab:lpips_alex} repeats it with AlexNet features under the same $[-1,1]$ input scaling, as in the evaluation code of TeaCache~\cite{liu2025teacache} and PAB~\cite{zhao2025pab}, leaving frames, pairs, and the per-video mean over frames unchanged.
Over all 4,250 video pairs behind the fidelity numbers of this paper, including the rank runs of Appendix~\ref{sec:equal_bytes}, the two backbones order the pairs almost identically (Spearman correlation 0.987), and the comparison of representations carries over: as with VGG features, INT4 stays closer to Native than LRQ-Cache in seven of the nine settings, and LRQ-Cache stays below SVD128 on 18 of the 50 HunyuanVideo TeaCache prompts, against 16 with VGG features (Fig.~\ref{fig:prompt_controls}).

\section{Complete results across three video models}
\label{sec:complete_results}
Table~\ref{tab:all_representations} lists every configuration of the three models: Dense inference, each native cache, and its INT4, SVD128, and LRQ-Cache representations.
Table~\ref{tab:main} and Table~\ref{tab:ablation} summarize the same runs; throughout the paper, ratios, differences, and relative changes are computed before rounding.

\begin{table*}[!t]
\centering
\caption{\textbf{All configurations across three video models.} Each row covers 50 inputs. Cache is complete persistent storage relative to the same native cache. Shaded rows are LRQ-Cache (ours). /N compares with the same native cache and /D with Dense inference.}
\label{tab:all_representations}
\small
\setlength{\tabcolsep}{4pt}
\begin{tabular}{@{}llrrrrrr@{}}
\toprule
Cache rule & Representation & Cache & VBench & PSNR/N & SSIM/N & LPIPS/N & LPIPS/D \\
\midrule
\multicolumn{8}{l}{\textbf{HunyuanVideo}} \\
Dense & Dense & -- & 84.71\% & -- & -- & -- & 0.0000 \\
TeaCache & Native & 100.00\% & 84.50\% & $\infty$ & 1.0000 & 0.0000 & 0.2217 \\
TeaCache & INT4 & 76.04\% & 84.50\% & 41.78 & 0.9883 & 0.0175 & 0.2216 \\
TeaCache & SVD128 (BF16) & 68.09\% & 84.27\% & 35.41 & 0.9701 & 0.0490 & 0.2281 \\
\rowcolor{oursbg} TeaCache & \textbf{LRQ-Cache} (ours) & \textbf{67.42\%} & 84.26\% & 35.37 & 0.9697 & 0.0493 & 0.2281 \\
TaylorSeer & Native & 100.00\% & 80.10\% & $\infty$ & 1.0000 & 0.0000 & 0.5508 \\
TaylorSeer & INT4 & 28.21\% & 79.96\% & 38.45 & 0.9637 & 0.0884 & 0.5564 \\
TaylorSeer & SVD128 (BF16) & 4.38\% & 79.02\% & 37.19 & 0.9651 & 0.0917 & 0.5453 \\
\rowcolor{oursbg} TaylorSeer & \textbf{LRQ-Cache} (ours) & \textbf{2.38\%} & 79.02\% & 37.19 & 0.9656 & 0.0911 & 0.5452 \\
ScalingCache & Native & 100.00\% & 84.65\% & $\infty$ & 1.0000 & 0.0000 & 0.0596 \\
ScalingCache & INT4 & 28.13\% & 84.67\% & 42.88 & 0.9920 & 0.0158 & 0.0571 \\
ScalingCache & SVD128 (BF16) & 4.33\% & 84.55\% & 38.42 & 0.9843 & 0.0331 & 0.0605 \\
\rowcolor{oursbg} ScalingCache & \textbf{LRQ-Cache} (ours) & \textbf{2.30\%} & 84.54\% & 38.23 & 0.9831 & 0.0345 & 0.0606 \\
\midrule
\multicolumn{8}{l}{\textbf{Wan2.1 T2V}} \\
Dense & Dense & -- & 84.82\% & -- & -- & -- & 0.0000 \\
TeaCache & Native & 100.00\% & 84.70\% & $\infty$ & 1.0000 & 0.0000 & 0.0949 \\
TeaCache & INT4 & 28.15\% & 84.69\% & 33.04 & 0.9666 & 0.0409 & 0.0960 \\
TeaCache & SVD128 (BF16) & 2.93\% & 84.62\% & 29.29 & 0.9344 & 0.0780 & 0.1168 \\
\rowcolor{oursbg} TeaCache & \textbf{LRQ-Cache} (ours) & \textbf{1.57\%} & 84.65\% & 29.28 & 0.9336 & 0.0784 & 0.1161 \\
TaylorSeer & Native & 100.00\% & 79.17\% & $\infty$ & 1.0000 & 0.0000 & 0.5918 \\
TaylorSeer & INT4 & 28.13\% & 74.62\% & 21.75 & 0.6633 & 0.3761 & 0.6343 \\
TaylorSeer & SVD128 (BF16) & 2.89\% & 76.10\% & 21.65 & 0.6857 & 0.3444 & 0.6158 \\
\rowcolor{oursbg} TaylorSeer & \textbf{LRQ-Cache} (ours) & \textbf{1.54\%} & 76.04\% & 21.65 & 0.6863 & 0.3450 & 0.6160 \\
ScalingCache & Native & 100.00\% & 84.62\% & $\infty$ & 1.0000 & 0.0000 & 0.1128 \\
ScalingCache & INT4 & 28.13\% & 83.07\% & 28.53 & 0.8984 & 0.1480 & 0.1826 \\
ScalingCache & SVD128 (BF16) & 2.89\% & 81.81\% & 24.90 & 0.8114 & 0.2790 & 0.2893 \\
\rowcolor{oursbg} ScalingCache & \textbf{LRQ-Cache} (ours) & \textbf{1.54\%} & 81.84\% & 24.88 & 0.8110 & 0.2794 & 0.2898 \\
\midrule
\multicolumn{8}{l}{\textbf{Wan2.1 I2V}} \\
Dense & Dense & -- & 84.42\% & -- & -- & -- & 0.0000 \\
TeaCache & Native & 100.00\% & 84.39\% & $\infty$ & 1.0000 & 0.0000 & 0.0623 \\
TeaCache & INT4 & 28.14\% & 84.40\% & 32.63 & 0.9565 & 0.0507 & 0.0626 \\
TeaCache & SVD128 (BF16) & 2.68\% & 84.44\% & 31.09 & 0.9444 & 0.0621 & 0.0706 \\
\rowcolor{oursbg} TeaCache & \textbf{LRQ-Cache} (ours) & \textbf{1.43\%} & 84.43\% & 31.16 & 0.9451 & 0.0616 & 0.0707 \\
TaylorSeer & Native & 100.00\% & 81.70\% & $\infty$ & 1.0000 & 0.0000 & 0.4137 \\
TaylorSeer & INT4 & 28.13\% & 77.33\% & 19.50 & 0.5157 & 0.5399 & 0.6055 \\
TaylorSeer & SVD128 (BF16) & 2.67\% & 78.23\% & 20.63 & 0.5971 & 0.4867 & 0.5666 \\
\rowcolor{oursbg} TaylorSeer & \textbf{LRQ-Cache} (ours) & \textbf{1.42\%} & 78.18\% & 20.63 & 0.5958 & 0.4883 & 0.5669 \\
ScalingCache & Native & 100.00\% & 84.41\% & $\infty$ & 1.0000 & 0.0000 & 0.0813 \\
ScalingCache & INT4 & 28.13\% & 83.71\% & 28.95 & 0.9011 & 0.1528 & 0.1658 \\
ScalingCache & SVD128 (BF16) & 2.67\% & 82.95\% & 25.65 & 0.8152 & 0.2839 & 0.2877 \\
\rowcolor{oursbg} ScalingCache & \textbf{LRQ-Cache} (ours) & \textbf{1.42\%} & 82.97\% & 25.63 & 0.8162 & 0.2838 & 0.2877 \\
\bottomrule
\end{tabular}
\end{table*}

\section{Paired prompt-level ablation analysis}
\label{sec:prompt_analysis}
\begin{figure*}[!t]
    \centering
    \includegraphics[width=\textwidth]{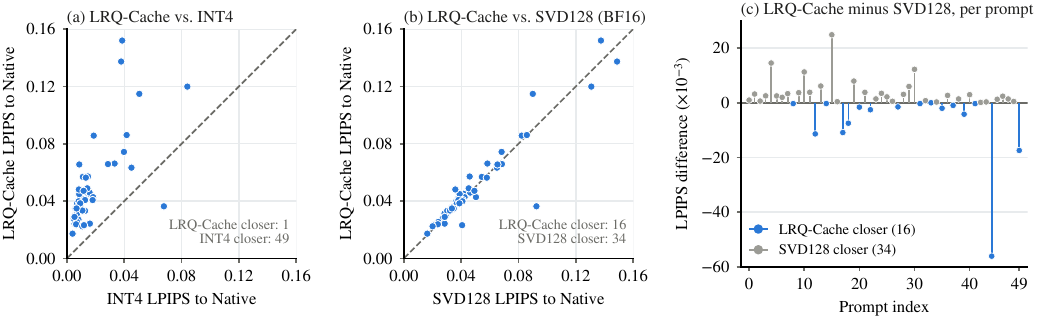}
    \caption{\textbf{Every-prompt comparison under HunyuanVideo TeaCache.} (a) LRQ-Cache has lower LPIPS to Native than INT4 on 1 of the 50 prompts and higher on 49.
    (b) Its distance to Native is lower than that of SVD128 on 16 prompts and higher on 34.
    Dashed diagonals denote equal scores.
    (c) The signed change for each prompt, blue where LRQ-Cache is closer to Native, shows the spread behind the small mean difference.
    These are differences between distances to a shared reference, not direct distances between LRQ-Cache and SVD128 outputs.
    Panels (a) and (b) use the same axis ranges.}
    \label{fig:prompt_controls}
\end{figure*}

The analysis reuses the whole-video scores behind Table~\ref{tab:all_representations} for all 50 HunyuanVideo TeaCache prompts at seed 0 and generates or scores no video of its own.
Each representation is compared with the uncompressed TeaCache output of the same prompt, and Fig.~\ref{fig:prompt_controls} covers every prompt, complementing the selected visual examples of the main paper.

\paragraph{Comparison with INT4.}
INT4 is closer to Native than LRQ-Cache on PSNR, SSIM, and LPIPS for 49 of the 50 prompts, so the ordering holds across the prompt set rather than in a selected example, although INT4 also stores a larger feature payload (28.13\% of native against 2.27\%, or 76.04\% and 67.42\% once the uncompressed probe is included).
This ranking is specific to the mechanism and budget examined here; Appendix~\ref{sec:equal_bytes} compares the two representations across storage budgets on Wan2.1 T2V.

\paragraph{Comparison at fixed rank.}
Table~\ref{tab:paired_precision} reports LRQ-Cache minus SVD128 differences, so a positive PSNR or SSIM difference favors LRQ-Cache, as does a negative LPIPS difference.
We resample the 50 paired prompt records with replacement 10,000 times (analysis seed 20261102), keep the within-prompt pairing, and take the 2.5th and 97.5th percentiles of each mean difference; the intervals describe prompt-resampling variability within the observed set, not variation over generation seeds.

\begin{table*}[!t]
    \centering
    \caption{\textbf{Paired factor-precision differences on 50 prompts.} The two arms use rank 128.
    Intervals are percentile intervals for the mean difference; the last column shows the empirical 95th percentile of the absolute per-prompt difference.}
    \label{tab:paired_precision}
    \small
    \begin{tabular}{lrrrr}
        \toprule
        Metric & Mean & Median & 95\% resampling interval & 95th percentile $|\Delta|$ \\
        \midrule
        PSNR (dB) & $-0.04$ & $-0.23$ & $[-0.43,\;0.47]$ & $2.19$ \\
        SSIM & $-0.0004$ & $-0.0003$ & $[-0.0030,\;0.0026]$ & $0.0209$ \\
        LPIPS & $+0.0003$ & $+0.0011$ & $[-0.0029,\;0.0028]$ & $0.0161$ \\
        \bottomrule
    \end{tabular}
\end{table*}

The near-zero mean LPIPS difference coexists with changes from $-0.0561$ to $+0.0248$ across prompts: an interval containing zero establishes no equivalence and no noninferiority margin was fixed in advance, so the data support a small average change at fixed rank with measurable per-prompt variation, and end-to-end differences can also reflect adaptive decisions taken later.

\section{Storage--quality trade-off and rank}
\label{sec:equal_bytes}
On Wan2.1 T2V with TeaCache, we vary the rank of LRQ-Cache over 32, 64, 128, 256, 512, 1024, and 2048 with INT8 factors, and the bit width of the token-group quantizer of INT4 (Sec.~\ref{sec:protocol}) over 2, 4, and 8 bits.
Every configuration uses the 50 prompts, seed, and metric code of Table~\ref{tab:main}; rank 128 and 4 bits are the LRQ-Cache and INT4 runs of the main tables, and rerunning both reproduced their videos bit for bit on a check prompt.
This sweep is a post-hoc analysis; the main results keep the rank fixed in advance.
No configuration changes the TeaCache full-step set on any prompt, so the comparison isolates the stored representation.

\begin{figure*}[!t]
    \centering
    \includegraphics[width=\textwidth]{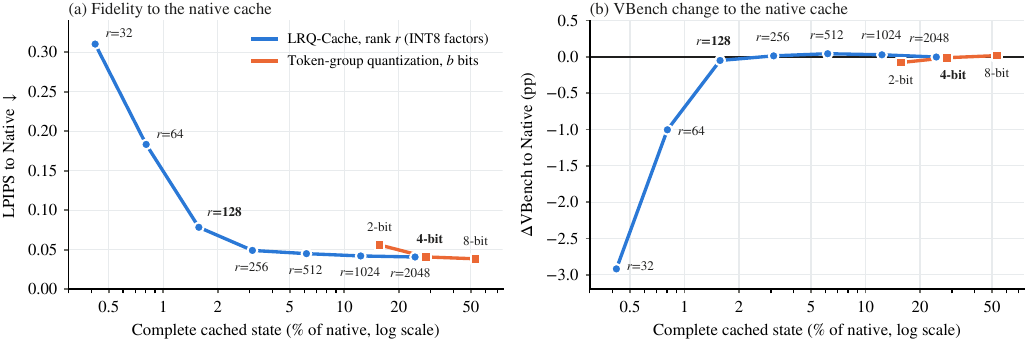}
    \caption{\textbf{Storage--quality trade-off on Wan2.1 T2V with TeaCache.} LRQ-Cache with rank $r$ and INT8 factors (blue) and token-group quantization of the dense tensor with $b$-bit codes (orange), against the complete cached state as a percentage of native (log scale), averaged over 50 prompts.
    (a) LPIPS to the same native cache.
    (b) VBench change relative to the native cache, in pp.
    Bold labels mark the configurations of the main tables: LRQ-Cache at rank 128 and INT4.}
    \label{fig:e1_pareto}
\end{figure*}

Fig.~\ref{fig:e1_pareto} plots fidelity and VBench against the complete cached state.
Rank 128 sits at the knee: ranks 256 and 1024 are closer to Native than 2-bit codes at a fraction of their state while rank 128 is not, the curve flattens above rank 256, and both metrics fall sharply below it.
Encoding takes 1.6\,s per video at rank 128 and 29.9\,s at rank 2048, against less than 0.5\,s for dense quantization, on shared hosts.

\paragraph{Rank on the predictive caches.}
On Wan2.1 T2V, we rerun LRQ-Cache at ranks 256 and 512 with both predictive caches, each fixed before its run, with the same codec, 50 prompts, seed, and metric code; for each cache, a rerun at rank 128 reproduced the main-table video bit for bit on a check prompt.
Table~\ref{tab:rank_taylor} lists the results.
TaylorSeer computes the same full steps in every run, each doubling of the rank doubles the state and lowers LPIPS to Native by 13--14\%, VBench recovers 1.27 and then 0.66\,pp, and all three ranks stay closer to Native than INT4 at under a quarter of its state; with ScalingCache the two higher ranks fall below INT4 on 45 of the 50 inputs and reduce the VBench loss from 2.78 to 1.34 and 0.64\,pp.
Its refresh rule reads the cached features, so the rank also changes which steps are full, as described in Appendix~\ref{sec:setup_details}.
These runs follow the main results and do not change their rank.

\begin{table*}[!t]
\centering
\caption{\textbf{Rank on Wan2.1 T2V predictive caches.} 50 prompts per row. State is the complete cached state as a percentage of native (150\,GiB); LPIPS (VGG, and AlexNet in parentheses) and PSNR compare with the same native cache, LPIPS/D with Dense, and $\Delta$VBench is the VBench change to the native cache in pp. Rank 128 and INT4 are the runs of Tables~\ref{tab:main} and~\ref{tab:ablation}.}
\label{tab:rank_taylor}
\small
\begin{tabular}{@{}lrrrrr@{}}
\toprule
Representation & State & LPIPS/N$\downarrow$ & PSNR/N$\uparrow$ & LPIPS/D$\downarrow$ & $\Delta$VBench$\uparrow$ \\
\midrule
TaylorSeer & 100\% & -- & -- & 0.5918 & -- \\
\quad + INT4 & 28.13\% & 0.3761 (0.2556) & 21.75 & 0.6343 & $-$4.54 \\
\quad + LRQ-Cache, rank 128 & 1.54\% & 0.3450 (0.2212) & 21.65 & 0.6160 & $-$3.12 \\
\quad + LRQ-Cache, rank 256 & 3.07\% & 0.2970 (0.1775) & 22.83 & 0.6081 & $-$1.85 \\
\quad + LRQ-Cache, rank 512 & 6.14\% & \textbf{0.2581} (\textbf{0.1457}) & \textbf{23.82} & \textbf{0.6024} & \textbf{$-$1.19} \\
\midrule
ScalingCache & 100\% & -- & -- & 0.1128 & -- \\
\quad + INT4 & 28.13\% & 0.1480 (0.0873) & 28.53 & 0.1826 & $-$1.55 \\
\quad + LRQ-Cache, rank 128 & 1.54\% & 0.2794 (0.1622) & 24.88 & 0.2898 & $-$2.78 \\
\quad + LRQ-Cache, rank 256 & 3.07\% & 0.1835 (0.0939) & 27.21 & 0.2109 & $-$1.34 \\
\quad + LRQ-Cache, rank 512 & 6.14\% & \textbf{0.1125} (\textbf{0.0539}) & \textbf{29.56} & \textbf{0.1637} & \textbf{$-$0.64} \\
\bottomrule
\end{tabular}
\end{table*}

\section{Additional paired video comparison}
\label{sec:qualitative_appendix}
\begin{figure*}[!t]
    \centering
    \includegraphics[width=\textwidth]{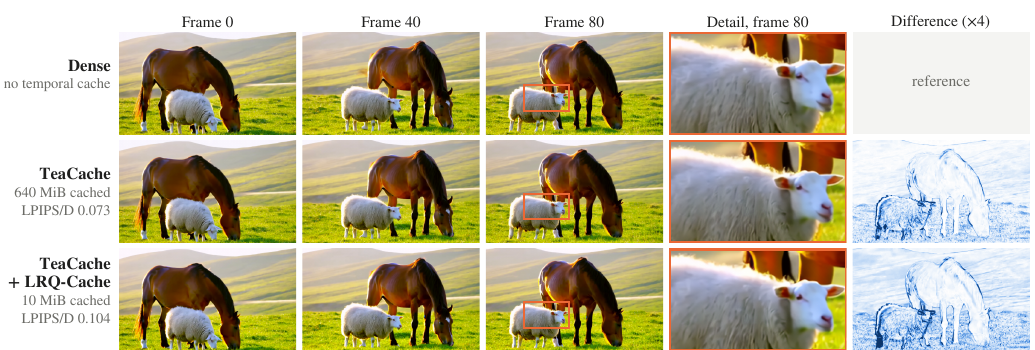}
    \caption{\textbf{Paired frames of Wan2.1 T2V with TeaCache.}
    Rows show Dense inference, TeaCache with its native cache, and TeaCache with LRQ-Cache, for one input chosen for coherent motion in the native video.
    The last column is the mean absolute RGB difference to Dense at frame 80, magnified four times, and LPIPS/D is the whole-video LPIPS to Dense.}
    \label{fig:qualitative_wan}
\end{figure*}

Figure~\ref{fig:qualitative_wan} repeats the layout of Fig.~\ref{fig:qualitative_primary} for a second model and mechanism, Wan2.1 T2V with TeaCache.
Prompt text, seed, and frame indices are identical within each comparison, and the detail crop uses the same source pixels in every row of Figs.~\ref{fig:qualitative_primary} and~\ref{fig:qualitative_wan}.
The complete videos behind both figures, their frame indices and crop coordinates, and further paired videos are included in the supplementary archive and described in its README.

\section{Quality dimensions}
\label{sec:quality_profile_appendix}
\begin{figure*}[!t]
    \centering
    \includegraphics[width=\textwidth]{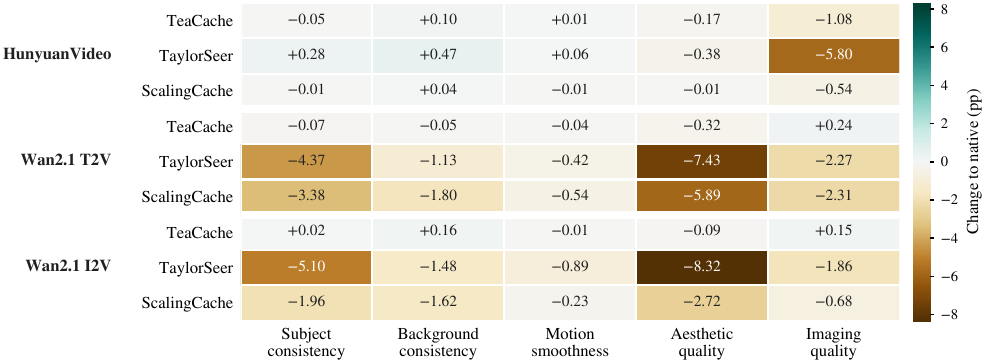}
    \caption{\textbf{Per-dimension quality changes relative to the same native cache.} Cells show $100\times(\mathrm{score}_{\rm LRQ}-\mathrm{score}_{\rm native})$ for each VBench dimension; positive values indicate increases and negative values decreases, in pp.
    Every reference is the same model and caching mechanism.}
    \label{fig:quality_dimensions}
\end{figure*}

Figure~\ref{fig:quality_dimensions} separates the five scores behind VBench.
On the Wan2.1 predictive caches, aesthetic quality and subject consistency change far more than motion smoothness, by $-7.43$ and $-4.37$\,pp on T2V and $-8.32$ and $-5.10$\,pp on I2V for TaylorSeer, against $-0.42$ and $-0.89$\,pp.
On HunyuanVideo TaylorSeer the loss sits in imaging quality (5.80\,pp) while the other four dimensions change by less than 0.5\,pp.
The aggregate mean hides these differences.

\section{Spectra of predictive caches}
\label{sec:structure_appendix}
\begin{figure*}[!t]
    \centering
    \includegraphics[width=\textwidth]{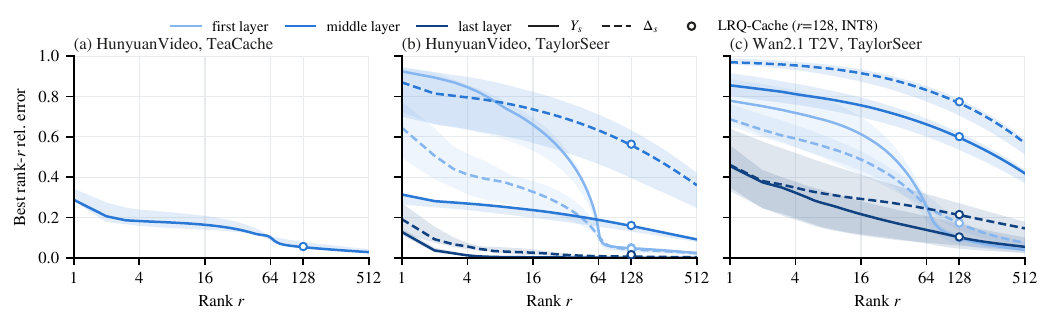}
    \caption{\textbf{Best rank-$r$ relative error of cached features at the main resolution.} Lines are medians over writes and bands their range; HunyuanVideo uses two evaluation prompts and Wan2.1 T2V one (\texttt{vbench-0319}, conditional branch, MLP outputs), each at three refreshes.
    Solid lines are module outputs $Y_s$ and dashed lines finite differences $\Delta_s$; circles mark LRQ-Cache with rank 128 and INT8 factors.
    Panels (a) and (b) show the data of Fig.~\ref{fig:tea_spectrum}. The rank was fixed before this measurement, and ScalingCache was not measured.}
    \label{fig:spectrum_all}
\end{figure*}

Figure~\ref{fig:spectrum_all} adds Wan2.1 T2V TaylorSeer to the two panels of Fig.~\ref{fig:tea_spectrum} and shows the range over writes, using the same main-resolution measurement.
At rank 128, the median error of the outputs of the first and last layers is at most 0.05 on HunyuanVideo and about 0.10 on Wan2.1 T2V, but the middle layers reach 0.16 and 0.56 for outputs and finite differences on HunyuanVideo and 0.60 and 0.77 on Wan2.1 T2V.
For every group of writes, the median error of the randomized INT8 factors of LRQ-Cache is within 0.015 of the best rank-128 error.

\section{Details of the analysis}
\label{sec:analysis_appendix}
This appendix derives the spectral and quantization bounds used in the paper and then accounts for storage and trajectory, applying standard low-rank approximation and norm inequalities to cached features.

\subsection{Shared features and low-rank approximation}
\label{sec:shared_analysis}
Let $X\in\mathbb R^{N\times D}$ be a nonzero cached feature matrix.
Define its mean feature $\mu$ and centered remainder $Z$ by
\begin{equation}
 \mu=\frac{1}{N}X^{\mathsf T}\mathbf1,
 \qquad Z=X-\mathbf1\mu^{\mathsf T}.
 \label{eq:mean_decomp}
\end{equation}
Here $\mathbf1$ is the vector of $N$ ones.
Every row of $\mathbf1\mu^{\mathsf T}$ is the same, so its rank is at most one, and $\mathbf1^{\mathsf T}Z=0$.
The shared and centered components are therefore orthogonal in the Frobenius inner product, giving
\begin{equation}
 \|X\|_F^2=N\|\mu\|_2^2+\|Z\|_F^2.
 \label{eq:shared_energy}
\end{equation}
The shared-energy fraction is $\rho=N\|\mu\|_2^2/\|X\|_F^2$.
It is a squared-energy statistic, not a fraction of the sum of singular values.
At the main resolution, its median over six cache writes is 0.914 (Sec.~\ref{sec:structural_argument}).
Because $\mathbf1^{\mathsf T}(Z_t-Z_s)=0$, the difference of two features splits into the same orthogonal components, $N\|\mu_t-\mu_s\|_2^2$ and $\|Z_t-Z_s\|_F^2$, so the shared fraction of the difference is smaller than that of $X_s$ whenever the relative change of the mean is smaller than that of the centered remainder.

For $1\leq r\leq\min(N,D)$, let $X_r$ be the optimal rank-$r$ approximation and $Z_{r-1}$ the optimal rank-$(r-1)$ approximation to $Z$.
Because $\mathbf1\mu^{\mathsf T}+Z_{r-1}$ has rank at most $r$,
\begin{equation}
 \begin{split}
 \|X-X_r\|_F^2
 &\leq\|X-(\mathbf1\mu^{\mathsf T}+Z_{r-1})\|_F^2\\
 &=\|Z-Z_{r-1}\|_F^2
 =\sum_{i\geq r}\sigma_i(Z)^2.
 \end{split}
 \label{eq:centered_tail}
\end{equation}
The first inequality follows from optimality, and the last equality is the standard truncated-SVD error formula.
Taking $r=1$ shows that the best rank-one approximation is no worse than the shared mean $\mathbf1\mu^{\mathsf T}$.
The bound concerns matrix approximation, not video quality.
A dominant shared component leaves the remaining rank budget available for centered variation.

\subsection{Low-rank error and factor quantization}
\label{sec:factor_analysis}
For an ideal truncated SVD, let $V$ contain the top $r$ right singular vectors and set $A=XV$, which equals $U_r\Sigma_r$ in Eq.~\ref{eq:svd}.
The approximation error is $E=X-AV^{\mathsf T}$, and the columns of $V$ are orthonormal, so $EV=0$.
Let the decoded factors be $A+\Delta A$ and $V+\Delta V$, and denote their Frobenius errors by $\delta_A=\|\Delta A\|_F$ and $\delta_V=\|\Delta V\|_F$.
In exact reconstruction arithmetic,
\begin{equation}
 \|X-\widehat X\|_F
 \leq\sqrt{\epsilon_r^2+\delta_A^2}
       +\|A+\Delta A\|_2\delta_V,
 \qquad
 \epsilon_r^2=\sum_{i>r}\sigma_i(X)^2.
 \label{eq:quant_bound}
\end{equation}
Expanding the reconstruction error gives
\begin{equation}
 X-\widehat X
 =E-\Delta A V^{\mathsf T}-(A+\Delta A)\Delta V^{\mathsf T}.
 \label{eq:factor_error_expansion}
\end{equation}
The first two terms are orthogonal because $EV=0$, and $\|\Delta A V^{\mathsf T}\|_F=\delta_A$ because $V^{\mathsf T}V=I$.
Their squared combined norm is therefore $\epsilon_r^2+\delta_A^2$.
Applying the triangle inequality and $\|PQ\|_F\leq\|P\|_2\|Q\|_F$ to the final term gives Eq.~\ref{eq:quant_bound}.
Coefficient-only quantization has the exact squared error $\epsilon_r^2+\delta_A^2$; basis quantization adds the final sensitivity term and need not preserve orthogonality.
Appendix~\ref{sec:factor_precision} measures the resulting error of the stored factors on recorded cache writes.

For a uniform quantizer without clipping, a group of $n$ values with scale $s$ and nearest rounding has squared error at most $ns^2/4$.
Summing over groups gives a bound for each factor.
If the stored scales are rounded, their effect belongs in the actual decoded perturbations $\Delta A$ and $\Delta V$.

The implementation uses randomized factorization rather than exact truncated SVD~\cite{halko2011finding}.
For arbitrary computed factors, the applicable general statement is
\begin{equation}
 \begin{split}
 \|X-\widehat X\|_F
 \leq{}& \|X-AV^{\mathsf T}\|_F
      +\delta_A\|V\|_2\\
      &+\|A+\Delta A\|_2\delta_V
      +\epsilon_{\rm arithmetic}.
 \end{split}
 \label{eq:practical_bound}
\end{equation}
The final term includes multiplication error and the cast to the original cache dtype.
The bound follows from the expansion and the triangle inequality, without requiring the computed factors to satisfy exact-SVD identities.

\subsection{Complete-cache storage}
\label{sec:storage_analysis}
\begin{figure*}[!t]
    \centering
    \begin{minipage}[t]{0.485\textwidth}
        \centering
        \includegraphics[width=\linewidth]{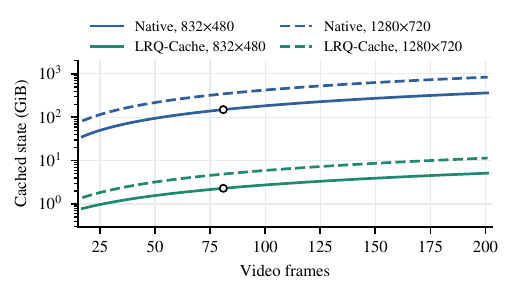}
        \caption{\textbf{Cached state of TaylorSeer on Wan2.1 T2V (14B).} Native state (Eq.~\ref{eq:state_bytes}) and LRQ-Cache packets (Eq.~\ref{eq:packet_bytes}, $r=128$) against video length at two resolutions; circles mark the measured states at $832\times480$ with 81 frames.}
        \label{fig:state_scaling}
    \end{minipage}\hfill
    \begin{minipage}[t]{0.485\textwidth}
        \centering
        \includegraphics[width=\linewidth]{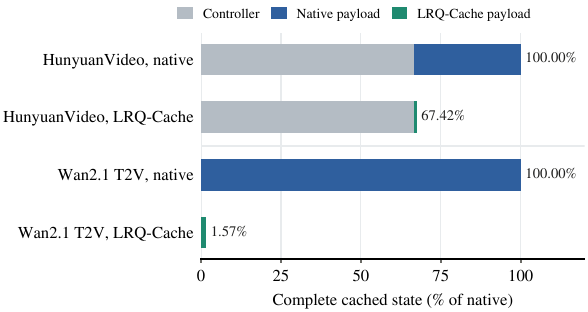}
        \caption{\textbf{Compressed payload versus complete cache.} Complete TeaCache states, normalized to the native state of each model. Gray denotes the retained controller tensors; the Wan2.1 T2V controller is included but is too small to resolve at this scale.}
        \label{fig:state_breakdown}
    \end{minipage}
\end{figure*}
Let $B_C$ be the native bytes of all features eligible for compression, $\widehat B_C$ their encoded bytes, and $B_U$ the bytes of controller or other unchanged tensors.
The fraction of complete persistent tensor storage that remains is
\begin{equation}
 \frac{\widehat B_C+B_U}{B_C+B_U}
 \geq\frac{B_U}{B_C+B_U}.
 \label{eq:whole}
\end{equation}
All quantities are nonnegative, so the right-hand side is the controller-only floor.
In the evaluated HunyuanVideo TeaCache layout, $B_C=729{,}907{,}200$ residual bytes and $B_U=1{,}459{,}814{,}400$ FP32 probe bytes, so $B_U=2B_C$ and the floor is $2/3$; the 16,575,488-byte packet leaves 67.42\% of the complete state.
Increasing compression of the residual alone cannot remove this state.
Wan2.1 T2V TeaCache retains a much smaller timestep probe: 670,924,800 residual bytes and 245,760 probe bytes with a 10,303,488-byte packet leave 1.57\%.

Encoded tensor bytes include integer codes, tensor-valued scales, and padding, with aliased storage counted once; non-tensor descriptors, allocator overhead, and decoded temporary tensors belong to separate accounting when the target is total resident or peak memory.
In particular, the large CPU-resident predictive states in our experiments are not GPU allocations of the same size.

\begin{figure*}[!t]
    \centering
    \includegraphics[width=\textwidth]{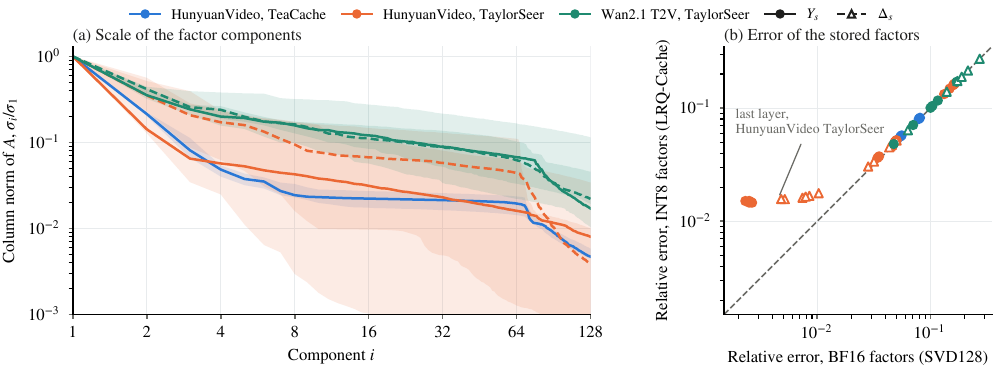}
    \caption{\textbf{Precision of the factors on the cache writes of Fig.~\ref{fig:spectrum_all}.} (a) Column norms of the FP32 factor $A$ relative to the first, $\sigma_i/\sigma_1$, for the 128 components; lines are medians over writes and bands their interquartile range. (b) Relative error $\|X-\widehat X\|_F/\|X\|_F$ of every write with INT8 factors (LRQ-Cache) against BF16 factors (SVD128) from the same randomized factorization; the dashed diagonal marks equal error. Solid lines and filled markers are module outputs $Y_s$, dashed lines and open markers finite differences $\Delta_s$.}
    \label{fig:factor_precision}
\end{figure*}

\subsection{How reconstruction error enters reuse}
\label{sec:readout_analysis}
At a TeaCache reuse step, the difference between the reconstructed and original cached residual enters the current addition with unit gain.
For a linear predictor $\sum_jw_jC_j$ with fixed coefficients, perturbations of the stored features satisfy
\begin{equation}
 \|\Delta\Psi\|_F\leq\sum_j|w_j|\,\|\Delta C_j\|_F.
 \label{eq:readout_gain}
\end{equation}
Eq.~\ref{eq:reuse_gain} is the case of two stored features with weights $1$ and $\kappa_t$; predictor weights can amplify feature errors.
The bound assumes fixed weights and applies locally; the full denoising process is not assumed linear.

For a conditional trajectory argument, let the state contain the video latent and the decoded logical cache/controller state.
Under a fixed trace of refresh and reuse steps, suppose the step map is locally Lipschitz with constants $L_t$ in the state and $K_t$ in a local representation perturbation of size $\varepsilon_t$.
If $e_t$ is the discrepancy between the two augmented states, then
\begin{equation}
 \begin{split}
 &e_{t+1}\leq L_te_t+K_t\varepsilon_t,\\
 &e_T\leq\sum_{t=0}^{T-1}K_t\varepsilon_t
                    \prod_{k=t+1}^{T-1}L_k
 \quad\text{when }e_0=0.
 \end{split}
 \label{eq:propagation}
\end{equation}
Repeated substitution gives the second inequality.
Perturbations include writes to the decoded cache state.
The constants are not estimated, so the bound is not a numerical certificate.
Adaptive branch changes fall outside this fixed-trace argument and are examined through closed-loop runs and recorded refresh traces.

\section{Precision of the factors}
\label{sec:factor_precision}
SVD128 stores the rank-128 factors of Eq.~\ref{eq:svd} in BF16; LRQ-Cache also quantizes them to 8-bit codes (Sec.~\ref{sec:factor_quant}), which halves the factor bytes and brings the complete predictive state to 2.30--2.38\% of native on HunyuanVideo and 1.42--1.54\% on Wan2.1 (Table~\ref{tab:all_representations}).
Figure~\ref{fig:factor_precision} shows why the halving costs little accuracy, on the 60 cache writes of Fig.~\ref{fig:spectrum_all}.

The column norms of $A$ are the singular values $\sigma_i$, and they fall steeply with the component index (Fig.~\ref{fig:factor_precision}a): the median ratio $\sigma_1/\sigma_{128}$ is between 46 and 257 for the five groups of writes.
A quantization range shared across components would therefore be set by the leading components, whereas the per-column groups of Sec.~\ref{sec:factor_quant} match the range of every component to its own scale.

With these groups, the INT8 factors stay close to the BF16 factors of the same randomized factorization (Fig.~\ref{fig:factor_precision}b): on 48 of the 60 writes the relative error rises by at most 8.1\% (median 0.4\%), and the other 12, the outputs and differences of the last HunyuanVideo TaylorSeer layer, are almost exactly of rank 128, where rounding dominates and the INT8 error stays within 0.018.

End to end the two representations agree in all nine settings: LPIPS to Native differs by at most 0.0016 and VBench by at most 0.06\,pp (Table~\ref{tab:all_representations}), and Appendix~\ref{sec:prompt_analysis} compares them prompt by prompt on HunyuanVideo TeaCache.
The cache writes cover only the mechanisms of Fig.~\ref{fig:spectrum_all}; for ScalingCache, Wan2.1 T2V TeaCache, and Wan2.1 I2V the comparison rests on these end-to-end results.


\end{document}